%% file: main.tex
\documentclass{article}

\usepackage{microtype}
\usepackage{graphicx}
\usepackage{subcaption}
\usepackage{booktabs}
\usepackage{hyperref}
\usepackage[accepted]{icml2026}
\usepackage{amsmath}
\usepackage{amssymb}
\usepackage{mathtools}
\usepackage{tikz}
\usetikzlibrary{arrows.meta,positioning}

\makeatletter
\renewcommand{\ICML@appearing}{Accepted at the 1st Workshop on Culture x AI:
Evaluating AI as a Cultural Technology, ICML 2026. Copyright 2026 by the author(s).}
\makeatother
\hypersetup{pdfsubject={Accepted at the 1st Workshop on Culture x AI: Evaluating AI as a Cultural Technology, ICML 2026}}

\icmltitlerunning{\smash{PAUSE}}

\begin{document}

\twocolumn[
\icmltitle{PAUSE: Editable Strategy Artifacts for Long-Form Cultural Story Adaptation}

\begin{icmlauthorlist}
\icmlauthor{Taaha Kazi}{pocketfm}
\icmlauthor{Vasu Sharma}{pocketfm}
\icmlauthor{Mohammad Saifullah}{pocketfm}
\icmlauthor{Abdur Rahman}{pocketfm}
\end{icmlauthorlist}

\icmlaffiliation{pocketfm}{Pocket FM}
\icmlcorrespondingauthor{Taaha Kazi}{taaha.kazi@pocketfm.com}

\icmlkeywords{cultural AI, narrative adaptation, transcreation, controllable generation, LLM-as-judge}

\vskip 0.3in
]

\printAffiliationsAndNotice{}

\input{sections/00_abstract}
\input{sections/F_mechanism_figure}
\input{sections/01_introduction}
\input{sections/02_related}
\input{sections/03_pipeline}
\input{sections/04_steering}
\input{sections/05_evaluation}
\input{sections/06_conclusion}

\bibliography{references}
\bibliographystyle{icml2026}

\input{sections/A_appendix}

\end{document}

%% file: sections/00_abstract.tex
\begin{abstract}
Generative AI systems increasingly mediate cultural adaptation, but
their cultural decisions are often hidden inside prompts, transient
model plans, or final prose. We study PAUSE
(\textbf{P}ause-\textbf{A}nd-\textbf{U}pdate \textbf{S}trategy
\textbf{E}diting), an intervention that exposes an editable adaptation
strategy as a human control surface for cultural decisions in long-form
story adaptation. The strategy is a structured artifact that can be
inspected, edited, and then projected through downstream character,
entity, and chapter-localization stages. In two Chinese-source serialized
novels, we test whether human edits to this strategy propagate into
chapter-level prose. Across $9$ edited-vs-control chapter comparisons,
judges select the edited-strategy output in all $9$; a marker audit
shows target markers in $8/9$ edited outputs and $0/9$ controls, with
forbidden markers absent from edited outputs and present in all controls.
We frame these results as a smoke-scale edit-adherence study, not a
claim that the outputs are culturally authoritative or literary-quality
improvements. PAUSE offers one practical way to make AI-mediated cultural adaptation
more inspectable and contestable before decisions propagate through
long-form generation.
\end{abstract}

%% file: sections/F_mechanism_figure.tex
\begin{figure*}[t]
\centering
\resizebox{0.96\textwidth}{!}{%
\begin{tikzpicture}[
font=\small,
box/.style={draw=black!50, rounded corners=2pt, align=center, minimum height=9mm, text width=24mm, fill=gray!8, inner sep=3pt},
strategy/.style={draw=orange!85!black, very thick, rounded corners=2pt, align=center, minimum height=12mm, text width=31mm, fill=orange!12, inner sep=3pt},
pathbox/.style={draw=black!35, rounded corners=2pt, align=center, minimum height=30mm, text width=46mm, fill=gray!4, inner sep=3pt},
pathlabel/.style={font=\scriptsize\bfseries, text=black!65},
directbox/.style={draw=orange!80!black, rounded corners=1.6pt, align=center, minimum height=5.2mm, text width=41mm, fill=orange!10, inner sep=1.4pt},
charbox/.style={draw=blue!70!black, rounded corners=1.6pt, align=center, minimum height=5.2mm, text width=41mm, fill=blue!7, inner sep=1.4pt},
entbox/.style={draw=green!45!black, rounded corners=1.6pt, align=center, minimum height=5.2mm, text width=41mm, fill=green!8, inner sep=1.4pt},
arrow/.style={-{Latex[length=2mm]}, line width=0.5pt, draw=black!65},
edit/.style={-{Latex[length=2mm]}, line width=0.6pt, draw=black!55, dashed}
]
\node[box] (src) at (0,0) {Source chapters\\$x_c$};
\node[box] (extract) at (2.8,0) {Extract +\\consolidate};
\node[strategy] (strategy) at (5.9,0) {Editable strategy\\$S'=E(S,\delta)$};
\node[pathbox] (paths) at (5.9,-2.35) {};
\node[box] (rewrite) at (9.95,-2.35) {Chapter rewrite\\$\pi(\cdot)$};
\node[box] (polish) at (12.70,-2.35) {Strategy-blind\\polish $\phi$};
\node[box] (final) at (15.30,-2.35) {Final output\\$y_c$};

\node[align=center, font=\small\bfseries] (human) at (5.9,1.40) {Human editor\\edit $\delta$};
\node[pathlabel] at (5.9,-1.35) {three propagation paths};
\node[directbox] at (5.9,-1.78) {Direct text $\rho(S')$};
\node[charbox] at (5.9,-2.35) {Character map $M_{\rm char}(S')$};
\node[entbox] at (5.9,-2.92) {Entity map $M_{\rm ent}(S')$};

\draw[arrow] (src.east) -- (extract.west);
\draw[arrow] (extract.east) -- (strategy.west);
\draw[edit] (human.south) -- (strategy.north);
\draw[arrow] (strategy.south) -- (paths.north);
\draw[arrow] (paths.east) -- (rewrite.west);
\draw[arrow] (rewrite.east) -- (polish.west);
\draw[arrow] (polish.east) -- (final.west);
\node[font=\scriptsize, text=black!60, align=center] at (12.70,-3.13) {no direct\\strategy input};
\end{tikzpicture}}
\caption{PAUSE mechanism. The pipeline captures an adaptation strategy $S$ from source chapters; a human edit $\delta$ produces $S'=E(S,\delta)$. The edited strategy reaches chapter rewrite through direct prompt text, character mappings, and entity mappings, while the strategy-blind polish pass receives only the rewritten draft.}
\label{fig:mechanism}
\end{figure*}
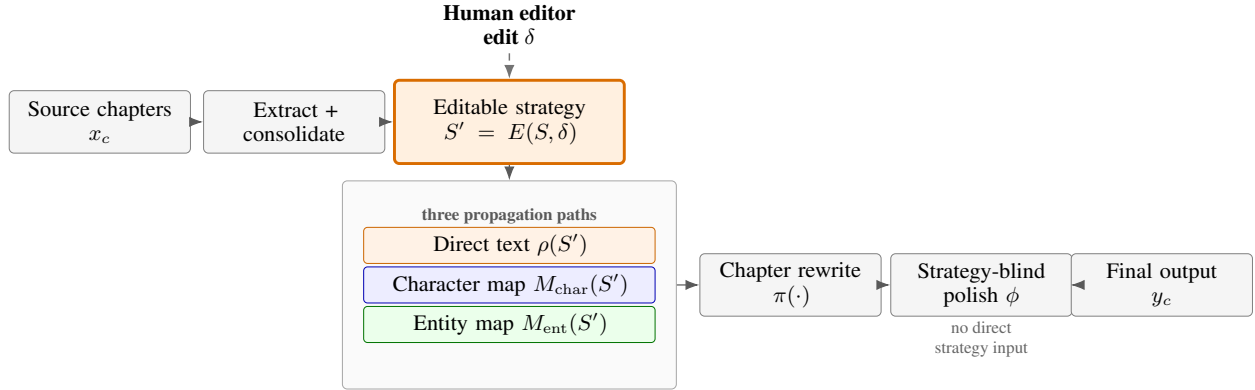

%% file: sections/01_introduction.tex
\section{Introduction}
\label{sec:intro}

Generative AI systems increasingly mediate cultural artifacts, but the
cultural decisions they make often remain hidden inside prompts,
transient model plans, or final prose. In long-form narrative
adaptation, these decisions become especially consequential because
names, settings, institutions, social hierarchies, and genre register
must remain stable across many
chapters~\citep{wang2023discobench,wang2023wmt23literary,wang2024wmt24literary,jin2024chapter2chapter}.
When the task is also cross-cultural, the problem is not only fluency:
translation may require transcreation, entity localization, and
context-sensitive cultural substitution~\citep{liu2024culturally,conia2024crosscultural}.

Current LLM workflows give human authors and editors two familiar
intervention points. They can steer through a prompt before generation
begins, hoping that the instruction survives the model's internal
planning and many downstream calls. Or they can post-edit the final
prose, after a global adaptation plan has already shaped names,
relationships, institutions, and style. Both points are useful, but
neither gives the editor a direct handle on the cultural decision layer
itself. For AI used as a cultural technology, this opacity matters:
cultural adaptation can erase source specificity, flatten cultures into
stereotypes, and automate parts of creative labor. The goal should not
be to make such transformations automatically correct, but to make them
inspectable and contestable before they propagate.

This paper studies PAUSE, a pause-and-update strategy editing
intervention built around an editable, interpretive intermediate
strategy. In our long-form adaptation pipeline, an LLM emits a structured
\emph{adaptation strategy} that captures the intended target
setting, naming conventions, institutional analogues, register choices,
and entity-localization principles. We expose this strategy as a JSON
artifact, allow a human editor to revise it, and resume the pipeline
from the edited copy without fine-tuning the model. The strategy is
	consumed by metadata-localization stages and projected into chapter
	writing through localized character records, entity records, mention
	mappings, and, in the patched experimental run, strategy-conditioned
	adaptation prompts; the final polish pass is deliberately
	strategy-blind. The empirical question is therefore
	not whether the system improves translation in general, but whether
	edits to this visible strategy artifact reach the prose that readers see.

We evaluate three edited-vs-control runs.\linebreak
Across the resulting $9$ chapter-level
pairwise comparisons from two Chinese-source web novels, judges select
the edited-strategy output in all
$9$. A deterministic marker audit points in the same direction: target
markers appear in $8/9$ edited outputs and $0/9$ controls, while
forbidden markers appear in $0/9$ edited outputs and $9/9$ controls.
	We treat this as a smoke-scale edit-adherence result. It shows that
	schema-aligned strategy edits can propagate into chapter prose; it does
	not establish cultural adequacy, literary quality, or full-novel
	coherence.

\paragraph{Contributions.}
(i)~We introduce PAUSE, identifying the LLM-generated adaptation strategy
as an editable control surface for cultural decisions in long-form
narrative adaptation. (ii)~We implement its pause/edit/resume mechanism
around this artifact without model fine-tuning. (iii)~We characterize three
propagation paths from strategy edits into prose: direct prose conditioning, indirect character mapping, and indirect entity mapping.
(iv)~We report a
smoke-scale propagation study over $9$ chapter comparisons, supported by
pairwise judgments and deterministic marker checks. (v)~We discuss the
limits of this evidence and the prerequisites for making cultural
adaptation workflows more inspectable and contestable.
Figure~\ref{fig:mechanism} gives the mechanism overview before we
separate the surrounding pipeline substrate from the edit-propagation
experiment.

%% file: sections/02_related.tex
\section{Related Work}\label{sec:related}

\paragraph{Long-form literary translation and cultural adaptation.}
Document- and chapter-level MT motivates our focus on decisions that
must remain consistent beyond a single sentence or scene~\citep{maruf2021survey}.
Recent WMT literary-translation shared tasks and the Disco-Bench benchmark
make discourse-level Chinese--English literary translation a concrete
evaluation setting~\citep{wang2023discobench,wang2023wmt23literary,wang2024wmt24literary};
\citet{jin2024chapter2chapter} further study chapter-to-chapter context.
A parallel line argues that translation across cultures cannot be reduced
to literal transfer: culturally aware NLP and cross-cultural MT require
attention to entities, social context, and target-culture
knowledge~\citep{liu2024culturally,conia2024crosscultural}. Our work
uses this setting, but asks a different question: whether the cultural
decision layer can be exposed as an editable artifact before it affects
many chapters of prose.

\paragraph{Human control through intermediate writing artifacts.}
Human-AI writing systems have often exposed control at the text-editing
or suggestion level, such as collaborative editors and interaction
datasets for creative writing~\citep{coenen2022wordcraft,lee2022coauthor}.
Long-form generation systems instead emphasize plans, outlines, memories,
and revision loops as intermediate objects that guide downstream
drafting~\citep{yao2019planwrite,yang2022re3,schick2022peer}. We follow
this intermediate-artifact pattern, but at a different granularity: the
edited object is not a local passage or plot outline, but a cross-chapter
adaptation strategy whose values are compiled into names, entities,
and chapter-localization prompts.

\paragraph{Evaluation of open-ended generation.}
Creative and cultural adaptation tasks rarely have one reference answer,
so pairwise and rubric-based LLM judgments are common but must be narrow
and bias-aware~\citep{zheng2023mtbench,liu2023geval}.
We therefore frame the judge task as edit adherence rather than global
quality, use judges from a different model family than the generator, and
add a deterministic marker audit. The marker audit follows the broader
evaluation pattern of decomposing long-form outputs into targeted,
checkable units, as in atomic factuality evaluation~\citep{min2023factscore}.
Together these choices position our evidence as a small propagation test,
not a benchmark for cultural quality or literary translation.

%% file: sections/03_pipeline.tex
\section{Pipeline Substrate}\label{sec:pipeline}

We describe only the parts of the long-form adaptation pipeline needed
to make the strategy-edit experiment legible. The pipeline separates a
\emph{metadata layer}, which extracts and localizes characters, entities,
relationships, and mention mappings, from a \emph{text layer}, which
uses those localized records to rewrite chapters. In the inspected
implementation these are separate entrypoints, but conceptually they are
one adaptation workflow. The model and judge details are reported with
the evaluation protocol.

\paragraph{Metadata layer}
For a source novel and target language/culture, the metadata layer runs:
(i)~per-chapter extraction of characters, entities, and relationships
into a knowledge graph (KG); (ii)~a strategy step that condenses the KG,
family structure, entity graph, and target setting into a structured
adaptation plan; (iii)~character localization; (iv)~entity localization;
and (v)~validation and cultural-status checks over the localized records.
The strategy is a JSON document containing declarative rules for names,
families, places, institutions, social address, genre terms, and cultural
references. Its schema is source-conditional: editors must revise the
keys the strategy LLM actually emitted, rather than assume a fixed
canonical schema.

\paragraph{Localized records and mention projection.}
Character localization produces canonical source-to-target name records,
including first names, surnames, honorifics, and mention maps. Entity
localization first handles locations, then localizes other entities by
graph segment so related institutions, places, objects, and cultural
items can be adapted together; graph segmentation uses Louvain-style
community detection ~\citep{blondel2008louvain}. After localization,
bridge scripts merge source and localized records and project them into
\texttt{mentions\_by\_chapter.json}, the machine-readable handoff from
metadata to chapter writing.

\paragraph{Text layer}
Chapter writing is a separate prototype entrypoint. It reads the source DOCX
and \texttt{mentions\_by\_chapter.json}, optionally augments mentions
from merged character/entity JSON, and writes each chapter in two passes.
Step~1 performs adaptation and mention replacement; in the patched
experimental run, relevant strategy text can also be injected into this
localize prompt. Step~2 rewrites the draft for fluency and style. The
polish pass is \emph{strategy-blind}: it does not see the strategy JSON
directly, so the evaluation tests whether strategy-sensitive decisions
introduced before polishing survive into the final chapter text.

Under PAUSE, we can summarize this claim boundary compactly. For source chapter
$x_c$, captured strategy $S$, human edit $\delta$, chapter rewrite
$\pi$, and strategy-blind polish $\phi$:
\begin{equation}
\begin{aligned}
S' &= E(S,\delta),\\
\mathcal{P}(S') &= \big(\rho(S'), M_{\rm char}(S'), M_{\rm ent}(S')\big),\\
\tilde{y}_c &= \pi\!\big(x_c,\mathcal{P}(S')\big),\\
y_c &= \phi(\tilde{y}_c).
\end{aligned}
\label{eq:propagation}
\end{equation}
Equation~\ref{eq:propagation} makes the claim boundary explicit:
$E$ is the edit operation, $\tilde{y}_c$ is the pre-polish chapter draft,
and $y_c$ is the final chapter output. $\mathcal{P}(S')$ is only a
compact notation for the three strategy-derived paths in
Figure~\ref{fig:mechanism}: $\rho(S')$ is direct strategy text injected
into the localize prompt, while $M_{\rm char}$ and
$M_{\rm ent}$ are the indirect character and entity mappings. A strategy
edit does not control every sentence; it can affect prose only through
these paths.

%% file: sections/04_steering.tex
\section{PAUSE: Editable Strategy Steering}\label{sec:steering}

PAUSE is deliberately small: expose the strategy artifact, edit it, and
resume from the edited copy. This places human input between
prompt-time steering and final-prose post-editing. The editor is not
asked to rewrite the whole output, and the model is not fine-tuned; the
editor changes the visible cultural decision layer that downstream
metadata-localization stages use.

\paragraph{Pause/edit/resume hook}
We implemented a local prototype hook around the generated strategy file.
In the first invocation, the pipeline runs through extraction and the
strategy LLM call, writes the strategy JSON to disk, and halts. The
editor revises that file. A second invocation
loads the edited JSON, continues metadata localization from that edited
strategy, and projects the resulting localized records into chapter
writing. The interface is therefore just the emitted schema: any text
editor can author an edit, but the edit must target fields that actually
exist in the captured strategy.

\paragraph{What is editable}
An edit is any change to the JSON document that the strategy step emits,
applied between the pause and the resume. In the family-musical-drama
case study (\emph{Love and Strings}), the editor reframes the
localization from a New York classical-piano family to a Nashville
country-guitar family. \textbf{One simple field edit} changes the
target-setting prose:
\begin{quote}
\small
\textbf{Before.}\ ``\ldots invoking the architecture, weather, and atmosphere of New York City and the broader New England area (e.g., brownstones, autumn foliage, the pace of Manhattan life).''
\\[2pt]
\textbf{After.}\ ``\ldots invoking the architecture, weather, and atmosphere of Middle Tennessee: rolling hills, limestone bedrock, hot summers, the neon glow of Lower Broadway, and the quiet elegance of Franklin's antebellum homes.''
\end{quote}
Other changes in the same edit affect surname conventions, institutions,
and honorifics. Appendix~\ref{app:edit-example} shows a longer
before/after excerpt. Whether any such edit actually reaches final prose
is empirical; Section~\ref{sec:eval} evaluates that propagation.

%% file: sections/05_evaluation.tex
\section{Evaluation}\label{sec:eval}
\suppressfloats[t]

We evaluate PAUSE through \emph{strategy-edit propagation}: whether a human edit to the
captured strategy changes the final chapter prose in the intended
direction. This is a smoke-scale study, not a full benchmark. The
controlled evidence covers two Chinese-source serialized novels,
\emph{Power of genes} and \emph{Love and Strings}, each with $99$ native
chapters, localized to American English.

\paragraph{Protocol}
Each comparison contains one source chapter, one control localization
resumed from the unedited captured strategy, one edited localization
resumed from the same captured strategy after the human edit, and an edit
description specifying target and forbidden phenomena. The two resumes
share the same source, captured baseline strategy, and downstream
configuration; the only intended input difference is the strategy edit.
A blinded pairwise judge sees the edit description and the two outputs,
then selects the output that better instantiates the edit while avoiding
forbidden remnants of the unedited strategy. This focused pairwise setup
follows common LLM-judge practice for open-ended outputs while keeping
the criterion to edit adherence rather than global
quality~\citep{liu2023geval,zheng2023mtbench}.
Because our judges are LLM-only, the pairwise results measure edit
adherence, not cultural authority or representativeness of target-culture
values.

\begin{table}[t]
\caption{Edited-vs-control propagation runs. G = \emph{Power of genes}; S = \emph{Love and Strings}.}
\label{tab:runs}
\centering
\scriptsize
\setlength{\tabcolsep}{1.8pt}
\begin{tabular}{@{}p{0.14\linewidth}p{0.59\linewidth}cc@{}}
\toprule
Case & Edit & Ch. & Wins \\
\midrule
G-2 &
Denver, CO $\rightarrow$ Oakdale, IL & 2 & 2/2 \\
G-99 &
Pittsburgh, PA $\rightarrow$ Oakdale, IL & 5 & 5/5 \\
S-20 &
NYC/Ashford $\rightarrow$ Nashville/Branson & 2 & 2/2 \\
\midrule
Total & & 9 & 9/9 \\
\bottomrule
\end{tabular}
\end{table}

\paragraph{Pairwise result}
Table~\ref{tab:runs} summarizes the three runs. The first \emph{Power of
genes} run edited a Denver/Colorado setting into a fictional Midwestern
town, Oakdale, Illinois. The second re-captured the strategy from the
full $99$-chapter source and edited a Pittsburgh/Pennsylvania setting to
Oakdale. The \emph{Love and Strings} run changed the target setting and
music culture from a New York classical-piano family to a Nashville
country-guitar family. Across all $9$ judged chapters, judges selected
the edited-strategy output in all $9$. The result shows that
schema-aligned edits can reach chapter prose; it does not show that the
system solves cultural adaptation or literary quality.

\paragraph{Marker audit}
To reduce dependence on judge preference, we also count edit-specific
surface markers in the saved outputs. Per-chapter target and forbidden
token counts are computed by case-insensitive word-boundary regex match
against a fixed token list derived from each edit description, applied to
the localized chapter outputs. For
example, the Oakdale edits target tokens such as \emph{Oakdale} and
\emph{Illinois}, while forbidding baseline tokens such as \emph{Denver}
or \emph{Pittsburgh}. Table~\ref{tab:markers} gives the aggregate.

\begin{table}[t]
\caption{Marker audit over the $9$ judged localized chapter outputs.}
\label{tab:markers}
\centering
\small
\begin{tabular}{@{}lcc@{}}
\toprule
Side & Target present & Forbidden present \\
\midrule
Edited & 8/9 & 0/9 \\
Control & 0/9 & 9/9 \\
\bottomrule
\end{tabular}
\end{table}

The one edited output without a target marker is an interior testing-room
scene that names neither city nor institution, so we treat it as a
no-op opportunity rather than a contradicted edit. The forbidden-marker
separation is complete: no edited output retains the forbidden baseline
markers, while every control output contains at least one.

\paragraph{Secondary rubric audit}
As a secondary guardrail, we re-judged all $9$ chapter pairs with
Opus~4.7 using shuffled A/B ordering and a four-axis $1$--$5$ rubric.
This yields $18$ forced-choice judgments; the edited side is selected in
$18/18$, and all $9$ chapters have edited-side wins under both orderings.
These scores are transcribed into the aggregation script; raw
judge-response files are not part of the current reproducibility package.
Table~\ref{tab:rubric} shows that the main difference is edit adherence:
source faithfulness is tied, while target-culture naturalness and prose
quality are similar in this small LLM-judged sample. We treat this as a
secondary audit, not as evidence that the outputs are culturally correct.

\begin{table}[t]
\caption{Secondary rubric audit ($n{=}18$ order-shuffled judgments).}
\label{tab:rubric}
\centering
\scriptsize
\setlength{\tabcolsep}{3pt}
\begin{tabular}{@{}lrrr@{}}
\toprule
Axis & Edited & Control & $\Delta$ \\
\midrule
Edit adherence & 4.67 & 1.00 & +3.67 \\
Source faithfulness & 4.22 & 4.22 & +0.00 \\
Target-culture naturalness & 4.50 & 4.06 & +0.44 \\
Prose quality & 4.44 & 4.17 & +0.28 \\
\bottomrule
\end{tabular}
\end{table}

\paragraph{Run settings}
The reported strategy, localization, and adaptation calls used
Gemini~2.5~Pro as the main generation model; two pairwise runs were
judged by GPT-5 with high reasoning effort and the five-chapter run by
Opus~4.7. The strategy step and chapter rewrite use live model aliases
without a fixed seed. Chapter rewrite is therefore non-deterministic in
absolute output, but each edited/control comparison is resumed from the
same captured baseline strategy and processed with the same downstream
configuration. This controls for the captured strategy and run setup,
with residual stochasticity remaining a limitation.

%% file: sections/06_conclusion.tex
\section{Discussion and Limitations}
\label{sec:conclusion}

\paragraph{Discussion}
PAUSE does not merely make cultural adaptation ``automatically correct''; it makes it transparent, controllable, and verifiable. The system exposes a consequential decision layer that is visible, editable, and testable before it propagates through any long-form generation workflow. This is the core claim of this paper: the most useful human interface is not another prompt box or final-prose post-edit, but the model's own intermediate cultural plan presented as an inspectable control surface. Cultural adaptation carries risks: erasing source specificity, flattening cultures into stereotypes, and automating creative labor. The editable strategy directly addresses these risks by making transformations inspectable and contestable before they reach the prose stage.

\paragraph{Limitations and future directions}
Our evaluation covers two source novels and Chinese-to-American-English
localization. The natural next step is to widen the study to more stories and
more language-culture pairs (the mechanism is already in use internally across
more than a dozen pairs, e.g., English$\to$Hindi and Korean$\to$Spanish) and
to confirm that the edit-adherence signal generalizes.

\paragraph{Conclusion}

A pause-and-resume hook on the strategy step lets a human revise the model's
intermediate cultural plan before it propagates into chapter prose. We verify
this propagation under three orthogonal signals---a blinded pairwise judge, a
model-free surface-marker audit, and a multi-axis re-judge with ordering
shuffle. The result is a precise and practical control surface for the global
creative choices of long-form localization.

%% file: sections/A_appendix.tex
\appendix

\section{A worked example of an edit}\label{app:edit-example}

To make the Section~\ref{sec:steering} description concrete, we show a before/after excerpt from the family-musical-drama case study (\emph{Love and Strings}). The author reframes a New York classical-piano ``Ashford'' family as a Nashville country-guitar ``Branson'' family. The two strategy sections shown are consumed by downstream localization, allowing the edit to reach chapter writing without rewriting prompts.

\paragraph{Naming convention edit.}
\begin{quote}
\small
\textbf{Before.} ``All character names belonging to the primary narrative (the American setting) must be fully Americanized with common Anglo-Saxon, Latinate, or otherwise recognizable American names. The specific pronunciation quirk of the source surname (T\'an/Q\'in) will be discarded in favor of a clear, consistent American surname. \ldots A new, consistent naming scheme will be established (e.g., The `\textbf{Ashford}' family).''

\textit{Foreign-character rule:} ``Should any character be explicitly introduced as foreign (e.g., a guest conductor from Germany, an exchange student from Japan), their name will retain its original phonetic structure to serve its narrative function as an outsider.''
\\[4pt]
\textbf{After.} ``\ldots A new, consistent naming scheme will be established (e.g., The `\textbf{Branson}' family).''

\textit{Foreign-character rule:} ``Should any character be explicitly introduced as foreign (e.g., \textbf{a record producer from Sweden, a touring fiddler from Ireland}), their name will retain its original phonetic structure \ldots''
\end{quote}

\paragraph{Honorifics-and-titles edit.} The original strategy specifies a complete title-mapping rubric tied to the classical-music milieu; the edit rewrites the entries that carry milieu-specific connotations and leaves the rest unchanged.
\begin{quote}
\small
\textbf{Before.} ``\textit{Patriarchal Titles:} `Old Master' becomes `\textbf{The Maestro}' in public/professional contexts \ldots `Big Master/Current Head' becomes `\textbf{Mr.\ Ashford}'. \textit{Female Family Heads:} `Senior Aunt' becomes `Ms.\ Catherine' or `\textbf{Ms.\ Ashford}'. The matriarch's professional title (`Professor') is used where appropriate (`\textbf{Dr.\ Ashford}').''
\\[4pt]
\textbf{After.} ``\textit{Patriarchal Titles:} `Old Master' becomes `\textbf{The Legend}' in public/professional contexts \textbf{(echoing the reverence given to iconic guitarists and bandleaders in Nashville)} \ldots `Big Master/Current Head' becomes `\textbf{Mr.\ Branson}'. \textit{Female Family Heads:} `Senior Aunt' becomes `Ms.\ Catherine' or `\textbf{Ms.\ Branson}'. The matriarch's professional title (`Professor') is used where appropriate (`\textbf{Dr.\ Branson}').''
\end{quote}

\paragraph{What this illustrates.} Both edits are intentional interpretive choices stored as prose values under existing strategy keys. No new top-level keys are introduced; no rule grammar is invoked. The edited decisions are projected through character localization, entity localization, mention mappings, and the chapter rewrite; the final polish pass does not see the strategy directly. The corresponding chapter-level outputs that result from this edit are reported in Section~\ref{sec:eval}.